# AN ODDS RATIO BASED INFERENCE ENGINE

DAVID S. VAUGHAN
BRUCE M. PERRIN
ROBERT M. YADRICK
MCDONNELL DOUGLAS ASTRONAUTICS COMPANY

PETER D. HOLDEN
KARL G. KEMPF
MCDONNELL DOUGLAS RESEARCH LABORATORIES

## ABSTRACT

Expert systems applications that involve uncertain inference can be represented by a multidimensional contingency table. These tables offer a general approach to inferring with uncertain evidence, because they can embody any form of association between any number of pieces of evidence and conclusions. (Simpler models may be required, however, if the number of pieces of evidence bearing on a conclusion is large.)

This paper presents a method of using these tables to make uncertain inferences without assumptions of conditional independence among pieces of evidence or heuristic combining rules. As evidence is accumulated, new joint probabilities are calculated so as to maintain any dependencies among the pieces of evidence that are found in the contingency table. The new conditional probability of the conclusion is then calculated directly from these new joint probabilities and the conditional probabilities in the contingency table.

## INTRODUCTION

The information for expert systems applications that involve uncertain inference can be represented by a multidimensional contingency table which has a dimension for each piece of evidence and a dimension for the conclusion. Suppose, for example, that each piece of evidence and the conclusion can have two states, true (or present) and false (or absent). Then each cell in the contingency table contains the joint probability for the associated states of each piece of evidence and the conclusion.

Unfortunately, with this scheme a contingency table for 19 pieces of evidence and one conclusion will have over a million cells. As Shortliffe and Buchanan (1975) point out, it is not possible to estimate such large numbers of probabilities satisfactorily. Thus, attention has focused on ways to avoid estimating such large numbers of joint probabilities.

One solution is to make strong assumptions concerning conditional independence among pieces of evidence and conclusions and to develop heuristics for combining different pieces of evidence. This approach is used in MYCIN (Shortliffe and Buchanan, 1975), PROSPECTOR (Duda, Hart, Nilsson, Reboh, Slocum, and Sutherland, 1977; Duda, Gashnig, and Hart, 1979), AL/X (Reiter, 1980; Paterson, 1981), and Dependence Trees (Chow and Liu, 1968). This approach presents theoretical problems when its strong assumptions are not met.

Another way to avoid independent estimation of large numbers of probabilities is to represent contingency tables by models which are parsimonious, but which do not require strong *a priori* assumptions concerning conditional independence. Modeling approaches of this type include linear, logit, and probit models (Cohen and Cohen, 1975) and log-linear models (Bishop, Feinberg, and Holland, 1975). Cheeseman (1983) described an approach for maximum entropy estimation of parsimonious contingency table models. This approach permits a wide variety of different constraint types to be imposed on log-linear model parameters and can be used to develop models of even very large contingency tables. In general, approaches of this type do not require strong assumptions concerning the structure of dependencies among pieces of evidence and conclusions. Instead, actual problem data or experts' judgments are used to estimate model parameter values which reflect the important dependencies existing in the particular application.

Suppose that a contingency table or a model of such a table is available. Another issue concerns using such a table or model for probabilistic inference. Suppose k possible pieces of evidence exist (e1, e2, . . ., ek) along with a conclusion c. Suppose further that, for each piece of evidence, the state of that evidence (i.e., true or false) is either known for certain or no information is available. In these cases, inference involves computing the conditional probability of the conclusion given the evidence for which the states are known for certain. For example, if e1 and e2 are known to be true, the conditional probability may be computed as:

$$P(c|e1,e2) = P(c \& e1 \& e2)/P(e1 \& e2) \qquad [1]$$

The required joint probabilities may be obtained from the contingency table or model. Cheeseman (1983) presents computationally efficient methods of computing such probabilities.

In many expert system applications, states of some pieces of evidence may not be known for certain, although the probabilities of various states may be known to be different from the marginal probabilities of the states in the original contingency table. For example, suppose a particular piece of evidence, e1, is either true or false, with marginal (*a priori*) probabilities in the contingency table of .4 and .6, respectively. Suppose, also, that evidence e2 is either true or false with marginal probabilities of .7 and .3, respectively. Suppose now that an observation (e.g., a diagnostic test) suggests new marginal (*a posteriori*) probabilities for e1 of .8 and for e2 of .9. Then the required conditional probability is $P(c|P(e1) = .8 \& P(e2) = .9)$.

The purpose of the present paper is to present a method for computing these conditional probabilities; in other words, to make inferences from contingency table models when the current state of the evidence is itself uncertain. The approach presented here is related to that of Cheeseman (1983) in that it involves (at least implicitly) updating log-linear model parameters. The approach is completely general, in that it makes no conditional independence assumptions. It can be applied for evidence and conclusions which can take on many states (e.g., the evidence "color" might be in states "red", "green", "blue", etc.), and it requires no prior designation of contingency table dimensions as being evidence or conclusions.

UNCERTAIN INFERENCE METHOD

Consider the example given above involving two pieces of evidence. The conditional probability is given by:

$$P(c|P(e1) = .8 \ \& \ P(e2) = .9) = P(c|e1 \ \& \ e2)P'(e1 \ \& \ e2) + P(c|e1 \ \& \ \overline{e2})P'(e1 \ \& \ \overline{e2}) + P(c|\overline{e1} \ \& \ e2)P'(\overline{e1} \ \& \ e2) + P(c|\overline{e1} \ \& \ \overline{e2})P'(\overline{e1} \ \& \ \overline{e2}),$$ [2]

where the conditional proabilities may be obtained from the original contingency table model and where the P' values are the new joint probabilities. This equation is readily generalized to situations involving more pieces of evidence and to evidence which has more than two states or categories. Equation 2 cannot be used as given to compute the required conditional probability, however, because it requires knowledge of the new joint probabilities of the evidence. All that is available are the new probabilities for the evidence. These are typically probabilities for each piece of evidence considered alone, rather than joint probabilities.

If the two pieces of evidence are independent, the required joint probabilities are equal to the products of the new probabilities (e.g., $P'(e1 \ \& \ e2) = .8 \times .9$). Often, pieces of evidence are not independent of each other and independence should not be assumed. Information concerning such dependencies is available in the original contingency table or parsimonious model. The uncertain inference approach presented here uses information in the original table and the new probabilities to obtain the new joint probabilities for the evidence. These new joint probabilities are then used in equation 2 (or its generalizations) to compute the needed conditional probability.

In general, changes in lower-order probabilities of a multi-dimensional contingency table imply changes in higher-order marginal probabilities and conditional probabilities of the table. Thus, it is not possible to construct new joint probabilities which reflect the new probabilities and the original conditional probabilities for the evidence. Instead, the present approach involves maintaining associations (as measured by odds ratios) among evidence from the original contingency table.

The odds ratio is useful for this purpose because it is a measure of association which is independent of particular marginal probabilities. Thus constant odds ratios can be maintained as probabilities are changed. For two pieces of evidence with new probabilities, one odds ratio is obtained from the original table and held constant:

$$\frac{\frac{P(e1\ \&\ e2)}{P(e1\ \&\ \overline{e2})}}{\frac{P(\overline{e1}\ \&\ e2)}{P(\overline{e1}\ \&\ \overline{e2})}} = \frac{P(e1\ \&\ e2)\ P(\overline{e1}\ \&\ \overline{e2})}{P(e1\ \&\ \overline{e2})\ P(\overline{e1}\ \&\ e2)} \qquad [3]$$

Consider now the contingency table for an inference network involving two pieces of evidence and one conclusion. Such a table is illustrated in Figure 1.

|  | C = False | | C = True | |
|---|---|---|---|---|
| E1 = False | P111 .05 | P121 .10 | P112 .20 | P122 .10 |
| E1 = True | P211 .10 | P221 .15 | P212 .25 | P222 .05 |
|  | E2 = False | E2 = True | E2 = False | E2 = True |

Figure 1: A contingency table for an inference network involving two pieces of evidence (E1 and E2) and one conclusion (c). Cell entries indicate joint probabilities for the indicated state of the evidence and conclusion (true or false).

In this case, four odds ratios are constrained to values from the original table as new joint probabilities are computed--a ratio for each pairwise association and a ratio reflecting the three-way association. For Figure 1, the pairwise odds ratio for e1 and e2 is given by:

$$\frac{P(111)\ P(121)\ P(212)\ P(222)}{P(211)\ P(221)\ P(112)\ P(122)} \qquad [4]$$

Likewise, the odds ratio for the three-way association is given by:

$$\frac{P(111)\ P(221)\ P(212)\ P(122)}{P(211)\ P(121)\ P(112)\ P(222)} \qquad [5]$$

This approach is easily generalized to any number of evidence pieces (Bishop, Feinberg, and Holland, 1975).

An example will illustrate the procedure described here. The contingency table of Figure 1 contains joint probabilities for two pieces of evidence (e1 and e2) and a conclusion (c). The marginal probabilities for e1 in this table are .45 (false) and .55 (true); the marginal probabilities for e2 are .6 (false) and .4 (true). Suppose that, for a particular problem, new probabilities for e1 are .3 (false) and .7 (true), and for e2 are .2 (false) and .8 (true). The first step is to obtain from the full contingency table the marginal contingency table for the evidence (in this case, e1 and e2). This subtable (illustrated in Figure 2A) is then transformed into a table whose marginals take on the new values and whose odds ratio(s) equal those in the original table. The new subtable is illustrated in Figure 2B.

(A) E1

|  | False | True |
|---|---|---|
| False | .25 | .20 |
| True | .35 | .20 |

False   True
E2

(B) E1

|  | False | True |
|---|---|---|
| False | .0493 | .2508 |
| True | .1507 | .5492 |

False   True
E2

Figure 2: Evidence subtables. Part A is the subtable from the full contingency table in Figure 1. Part B is the same table adjusted for new E1 and E2 probabilities of .70 and .80, respectively. Note that the odds ratio for the association between E1 and E2 has remained constant.

Finally, the joint probabilities from the new subtable are substituted into an appropriate version of equation 2 to obtain the needed conditional probability.

COMPUTING METHODS

This section considers methods for adjusting contingency tables in accordance with new probabilities while maintaining the original odds ratios. We currently use an algorithm known as iterative proportional fitting for this purpose. This algorithm has been used extensively to adjust contingency tables (Brown, 1959; Deming and Stephan, 1940; Bishop, Feinberg, and Holland, 1975).

In the present application, the algorithm starts with the original subtable for which new probabilities are available (e.g., Figure 2A).

Cell entries are multiplied iteratively by ratios of desired to current probability values for each set of marginals to be adjusted. All appropriate odds ratios in the starting table are maintained automatically in the final table.

A possible disadvantage of this algorithm is that the entire subtable must be explicitly constructed and stored. However, several factors should mitigate this potential problem. Even in an expert system that involves large numbers of evidence, only subsets of the evidence are relevant to any particular conclusion. Furthermore, uncertain information is likely to be available for only a subset of the evidence related to a particular conclusion. For the remaining evidence, information is certain (i.e., the evidence is known to be true or false) or no information is available. Thus, the sizes of subtables required in practice usually will be considerably smaller than is implied by the original full table.

Despite the reasons stated above, situations may arise in which the iterative proportion fitting algorithm cannot be used due to the size of the subtables. Unfortunately, one cannot predict which associations will or will not exist in a marginal subtable, even if one has a parsimonious model of the complete contingency table. This is because indirect effects in the full table may be reflected as associations between pieces of evidence in the marginal subtable. For example, if two pieces of evidence are both associated with a conclusion in a complete table, but are not associated with each other, they may still have a non-null odds ratio in the subtable. As a consequence, one can fit parsimonious models only if one assumes that certain effects are null (corresponding odds ratios are equal to unity) or if one fits and compares several models. Circumstances may arise in which it will be preferable to assume that higher-order conditional dependencies are absent rather than to assume that no dependencies exist among pieces of evidence. In cases such as this, Cheeseman's (1983) maximum entropy model estimation procedure would be very useful. Cheeseman's procedure also would be useful in cases for which new marginal probabilities for evidence are in the form of intervals rather than new point values.

SUMMARY

The examples in this paper have dealt with only the simplest kinds of inference networks. The approach, however, is much more general. It can be extended to instances of multi-category evidence, and to models that involve any form of association (conjunctive, disjunctive, or independent) between any number of pieces of evidence and conclusions. No other uncertain inference scheme appears to offer this flexibility while avoiding the theoretical problems inherent in the strong assumption of conditional independence.

Future research is necessary concerning the practical value of the present approach compared with other approaches for using contingency tables for uncertain inference, such as that of Cheeseman (1983). Additional effort will also be required to improve the computational efficiency of the present method for use in a full scale implementation.

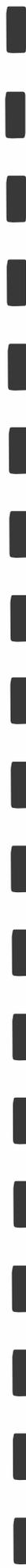